\documentclass[conference]{IEEEtran}
\IEEEoverridecommandlockouts

\usepackage{latexsym}
\usepackage[T1]{fontenc}
\usepackage[utf8]{inputenc}
\usepackage{microtype}
\usepackage{inconsolata}
\usepackage{graphicx}
\usepackage{cite}   

\usepackage{booktabs}
\usepackage{multirow}
\usepackage{amsmath}
\usepackage{amssymb}
\usepackage{xcolor}
\usepackage{colortbl}
\usepackage{enumitem}
\usepackage{xspace}
\usepackage{makecell}
\usepackage{tikz}
\usetikzlibrary{shapes.geometric, arrows.meta, positioning, decorations.pathreplacing, calc}
\usepackage{pifont}
\usepackage{stfloats} 

\newcommand{\etal}{\emph{et~al.}\xspace}

\makeatletter
\newcommand{\linebreakand}{%
  \end{@IEEEauthorhalign}
  \hfill\mbox{}\par
  \mbox{}\hfill\begin{@IEEEauthorhalign}
}
\makeatother

\begin{document}

\title{Entity Resolution in Practice:\\Lessons from a Self-Serve Pipeline}

\author{
\IEEEauthorblockN{Kaushik Pavani}
\IEEEauthorblockA{\texttt{kaushik.pavani@walmart.com}}
\and
\IEEEauthorblockN{Ganga Aluri}
\IEEEauthorblockA{\texttt{ganga.aluri@walmart.com}}
\and
\IEEEauthorblockN{Pravin Jadhav}
\IEEEauthorblockA{\texttt{pravin.jadhav@walmart.com}}
\linebreakand
\IEEEauthorblockN{Neeraj Prasad}
\IEEEauthorblockA{\texttt{neeraj.prasad@walmart.com}}
\and
\IEEEauthorblockN{Kiran Sanka}
\IEEEauthorblockA{\texttt{kiran.sanka@walmart.com}}
}

\maketitle

\begin{abstract}
We built and evaluated a self-serve entity resolution (ER) system on six
benchmarks spanning 864 to 5\,M records, and three lessons emerged that
are absent from existing ER literature.
\textbf{(1)~No single matching algorithm wins everywhere}---a self-serve
pipeline cannot predict its next dataset, so we recommend training several
algorithm families per dataset and letting an automatic bake-off pick the
winner.
\textbf{(2)~Precision and recall need separate fixes, not a shared
threshold}---precision needs hard rule-based vetoes, recall needs more
diverse candidate retrieval.
\textbf{(3)~One false-positive link can silently merge unrelated
entities}---assuming ``A matches B'' and ``B matches C'' implies ``A
matches C'' lets a single bad link chain hundreds of records together, so
every cross-group merge must be actively re-verified.
We hope these lessons save practitioners the months of dead-end
experiments that led us to them.
\end{abstract}


\section{Introduction}
\label{sec:intro}

\begin{figure*}[!b]
\centering
\small
\renewcommand{\arraystretch}{1.15}
\begin{tabular}{@{}llllll@{}}
\toprule
\textbf{ID} & \textbf{Name} & \textbf{Phone} & \textbf{Address} & \textbf{City} & \textbf{True} \\
\midrule
\rowcolor{blue!15}
R1 & Sakura Sushi     & 503-0147 & 42 Oak St & Portland & A \\
\rowcolor{blue!15}
R2 & Sakura Sushi Bar & 503-0147 & 42 Oak St & Portland & A \\
\midrule
\rowcolor{yellow!40}
R3 & Sakura Sushi & ---      & ---       & ---      & ? \\
\midrule
\rowcolor{red!12}
R4 & Sakura Sushi & 206-9283 & 8 Elm Ave & Seattle  & B \\
\rowcolor{red!12}
R5 & Sakura       & 206-9283 & 8 Elm Ave & Seattle  & B \\
\bottomrule
\end{tabular}
\vspace{4pt}

{\small \colorbox{blue!15}{\phantom{x}}~Entity A (Portland) \quad
        \colorbox{yellow!40}{\phantom{x}}~Bridge record (sparse) \quad
        \colorbox{red!12}{\phantom{x}}~Entity B (Seattle)}
\caption{\textbf{A preview of a typical failure mode in ER systems, and
one of the lessons we discuss in this paper.}
Consider the task of deduplicating restaurant records so that one cluster
represents one physical location (a single Sakura Sushi at 42~Oak~St in
Portland) rather than a brand across cities.
Five similarly named records arrive (R1--R5).
Most pairwise matchers will accept both R1$\leftrightarrow$R3 and
R3$\leftrightarrow$R4: the sparse bridge R3 has nothing in any populated
field that disagrees with either group.
The pipeline then chains the two correct-looking links together,
silently merging the Portland and Seattle locations into one cluster---an
incorrect answer under the per-location definition above.
We study this in \S\ref{sec:lesson3} and propose two fixes: (a)~a
\emph{verified-merge} clustering step, and (b)~a sparsity-aware confidence
threshold for records with few populated fields.}
\label{fig:bridge}
\end{figure*}

Entity resolution (ER)---identifying records that refer to the same real-world
entity---is foundational to data integration~\cite{christophides2021overview,
fellegi1969theory}.
Recent LLM-based approaches~\cite{peeters2025entity, fan2024cost} can match
records zero-shot but cost hundreds of dollars per million pairs and embed
matching logic in opaque weights.
PLM-based systems~\cite{li2020deep, mudgal2018deep} are cheap at inference
but require thousands of labeled pairs per domain and offer no audit trail.
Neither approach works for organizations running dozens of ER tasks under
changing requirements, strict auditability, and tight cost budgets.

We built a system to bridge this gap.
The key idea is simple: a structured YAML specification---a Standard
Operating Procedure (SOP)---encodes matching logic as inspectable, versionable
configuration.
An LLM teacher conditioned on the SOP labels candidate pairs; those labels
train a lightweight matcher via distillation at orders-of-magnitude lower
cost.
The SOP simultaneously prompts the teacher, structures its output, and serves
as the audit trail.

This paper is \emph{not} a systems paper.
It is a practitioner's guide organized around three failure modes we
encountered while evaluating this pipeline on six benchmarks spanning four
orders of magnitude (864 to 5\,M records)---failure modes absent from
existing ER literature:

\begin{enumerate}[leftmargin=*,nosep,topsep=2pt,label=\textbf{L\arabic*.}]
\item \textbf{No single matcher wins; let them compete} (\S\ref{sec:lesson1}).
A tournament over three canonical architectures (DeepMatcher, LightGBM, GAT)
auto-selects the best per dataset. Under a data-scarce regime
($\leq$10K training records), DeepMatcher and LightGBM each win 3/6
benchmarks on Pair-F1; GAT wins none.

\item \textbf{Precision and recall need separate toolkits}
(\S\ref{sec:lesson2}). Hard vetoes for precision, blocking ensemble
diversity for recall. No single threshold can optimize both.

\item \textbf{One false positive can collapse your clusters}
(\S\ref{sec:lesson3}). Transitive closure creates silent mega-clusters;
verified merge clustering with active cross-cluster verification recovers
recall safely.
\end{enumerate}

\section{The Framework}
\label{sec:framework}

Our system follows the standard \emph{block $\to$ match $\to$ cluster} ER
architecture~\cite{christophides2021overview, papadakis2020blocking}.
Three design choices motivate the lessons below.

\textbf{(1) Domain knowledge lives in an SOP, not weights.}
What counts as a match is a \emph{business decision}---e.g., two
food-court tenants share one phone number but are different
entities---and cannot be inferred from data without examples no
organization possesses.
We encode such rules in a versioned YAML SOP that serves three roles:
LLM teacher prompt, distillation signal (per-field similarity
assessments richer than a binary label), and audit trail.
A full SOP example is in Appendix~\ref{app:sop_yaml}.

\textbf{(2) Retrieval and matching are separate components.}
A \emph{blocker encoder} (Siamese fine-tuning, contrastive loss)
optimizes for recall; a \emph{matcher} optimizes for precision.
Training them separately avoids the tension inherent in a single
end-to-end model.
The matcher consumes blocker embeddings plus schema-driven features
(Jaro-Winkler, exact match, transposition detection) and is selected via
a \emph{tournament} over three canonical families
(Table~\ref{tab:matchers}).
The full pipeline---onboarding, training, and inference---is detailed in
Appendix~\ref{app:framework} with the architecture diagram
(Figure~\ref{fig:arch}) and model equations.

\textbf{(3) Per-dataset tuning is mandatory.}
The pipeline exposes $\sim$40 hyperparameters whose optimal values depend
on duplicate density, field sparsity, schema width, and scale, so we tune
per dataset.
For each experiment we obtain a strong baseline configuration using both
Optuna TPE Bayesian search~\cite{akiba2019optuna} and an LLM autoresearch
agent~\cite{karpathy2026autoresearch} and iterate from there; the
lessons below hold whichever search method produced the configuration.

\begin{table}[t]
\centering
\small
\caption{Matcher roster: one from each principal family.
\emph{Input}: E=embeddings, A=attribute features, G=graph structure.}
\label{tab:matchers}
\begin{tabular}{@{}l l c@{}}
\toprule
\textbf{Matcher} & \textbf{Family} & \textbf{Input} \\
\midrule
DeepMatcher & MLP~\cite{mudgal2018deep}      & E+A \\
LightGBM    & GBDT~\cite{ke2017lightgbm}     & A   \\
GAT         & GNN~\cite{brody2022attentive}  & E+G \\
\bottomrule
\end{tabular}
\end{table}

\section{Experimental Setup}
\label{sec:setup}

\paragraph{Datasets.}
We evaluate on six deduplication benchmarks spanning five domains
(Table~\ref{tab:datasets}) and four orders of magnitude in scale.
\emph{NCV} denotes the 5M-record benchmark of Saeedi
\etal~\cite{saeedi2018scalable}; it contains only generic structured fields used
for ER evaluation---no behavioral, financial, or sensitive attributes.
No proprietary, customer, or industry data is used anywhere; all experiments
are reproducible from the cited public benchmarks.

\begin{table}[t]
\centering
\small
\caption{Benchmark datasets.
$|\mathcal{R}|$=records, $|\mathcal{S}|$=schema fields,
$|\mathcal{C}|$=ground-truth clusters,
Sp.\ = fraction of pairs with missing fields.
Citations: \cite{naumann2010hpi,saeedi2017matching,saeedi2018comparative,saeedi2018scalable}.}
\label{tab:datasets}
\setlength{\tabcolsep}{3pt}
\begin{tabular}{@{}llrrrr@{}}
\toprule
\textbf{Dataset} & \textbf{Domain} & $|\mathcal{R}|$ & $|\mathcal{S}|$ & $|\mathcal{C}|$ & \textbf{Sp.} \\
\midrule
Restaurants & Restaur. &      864 &  5 &      752 & 2\%  \\
Cora        & Biblio.  &    1,879 & 17 &      182 & 68\% \\
Geo Settl.  & Geogr.   &    3,054 &  3 &      820 & 11\% \\
DBLP-Sch.   & Biblio.  &   66,879 &  4 &   61,604 & 5\%  \\
MB 200K     & Music    &  193,750 &  8 &  100,000 & 31\% \\
NCV         & Record   &5,000,000 &  4 &3,500,840 & 3\%  \\
\bottomrule
\end{tabular}
\end{table}

\paragraph{Splits and protocol.}
We split at the \emph{entity} level---no record from the same entity appears
in both training and test---and \emph{cap training and validation at 10K
records each}, regardless of dataset size, to reflect real-world deployment
where labeling requires domain expertise.
This yields heavily skewed ratios: Cora uses a conventional 42/13/45 split
(1.9K records), but MusicBrainz 200K trains on just 5\% (10K of 194K
records).
This design is deliberate: a method requiring abundant labels is impractical
for self-serve deployment.
We use a commercially available frontier LLM as teacher,
all-MiniLM-L6-v2 ($d{=}384$) as base encoder, and fixed seed 42.
\textbf{Pair-F1} is the primary metric throughout~\cite{mudgal2018deep};
purity is reported as a secondary metric to distinguish over-splitting from
over-merging.

\section{Lesson 1: Which Matcher Wins Depends on the Dataset}
\label{sec:lesson1}

\noindent\textbf{Claim.} No single matcher architecture dominates across ER
problems. A tournament that trains three canonical architectures and
auto-selects the winner eliminates a key human decision point.

\paragraph{Evidence.}
Table~\ref{tab:tournament} shows tournament results across all six
benchmarks. \emph{No single family dominates}, and the winning architecture
\emph{changes character} across datasets.

\begin{table}[t]
\centering
\small
\caption{Tournament leaderboard: Pair-F1 on held-out test
($\leq$10K training records). Winner in \textbf{bold}. Purity in
parentheses. $^\dagger$margin $<0.001$.}
\label{tab:tournament}
\setlength{\tabcolsep}{2.5pt}
\begin{tabular}{@{}l ccc l@{}}
\toprule
\textbf{Dataset} & \textbf{DM} & \textbf{LGBM} & \textbf{GAT} & \textbf{Winner} \\
\midrule
Restaur.\   & 0.948 {\tiny(.99)} & \textbf{0.969} {\tiny(1.0)} & 0.748 {\tiny(.99)} & LGBM \\
Cora        & \textbf{0.968} {\tiny(.98)} & 0.891 {\tiny(.98)} & 0.809 {\tiny(.89)} & DM \\
Geo Set.    & \textbf{0.979} {\tiny(.99)} & 0.960 {\tiny(.99)} & 0.964 {\tiny(.99)} & DM \\
DBLP-Sch.   & 0.160 {\tiny(1.0)} & \textbf{0.541} {\tiny(.94)} & 0.239 {\tiny(1.0)} & LGBM \\
MB 200K     & \textbf{0.964} {\tiny(1.0)} & 0.948 {\tiny(.99)} & 0.833 {\tiny(.95)} & DM \\
NCV         & 0.992 {\tiny(1.0)} & \textbf{0.993} {\tiny(1.0)} & 0.989 {\tiny(1.0)} & LGBM$^\dagger$ \\
\midrule
\multicolumn{5}{@{}l@{}}{\footnotesize \textbf{Score:} DM 3/6, LGBM 3/6, GAT 0/6.} \\
\bottomrule
\end{tabular}
\end{table}

\paragraph{Why the winner changes.}
Each winner reflects structural properties of its dataset.
\emph{DeepMatcher} wins on Cora, Geo Settlements, and MB 200K---datasets
where field-level attention and soft semantic similarity matter (sparse
attributes with OCR noise, paraphrase equivalence, subtle variant
spellings).
\emph{LightGBM} wins on Restaurants, DBLP-Scholar, and NCV---datasets that
are either small (the embedding tower lacks signal to fine-tune
meaningfully) or dominated by structured identifier fields where exact-match
and Jaro-Winkler features suffice.
\emph{GAT} wins nothing: at our 10K training cap, no dataset provides
enough connected-component structure for 2-hop graph attention to outperform
attribute-based methods, and GAT also suffers a train/test graph mismatch
when the $k$-NN graph at inference is built on a much larger test split.

\paragraph{Self-serve systems cannot pick in advance.}
The winning architecture changes with dataset size, schema sparsity, and
entity density---properties not known before running the data.
A fixed ``always DeepMatcher'' policy loses on Restaurants, DBLP-Scholar,
and NCV; ``always LightGBM'' loses on Cora, Geo Settlements, and MB 200K.
The tournament costs nothing extra---all three matchers share the same
training pairs and embeddings---and removes a decision point that would
otherwise require dataset-specific expertise.

\paragraph{Cost and latency.}
The teacher-student paradigm makes the tournament practical: the LLM teacher
labels once during training; the distilled matcher handles all inference.
The teacher costs $\sim$\$450/1M pairs at $\sim$2\,s per pair; the
tournament-winning matcher costs \$12/1M pairs---a $37.5\times$ cost
reduction.
LightGBM winners reach 222--263K pairs/sec on CPU; DeepMatcher winners run
at 5--10K pairs/sec including SBERT inference.

\paragraph{Practitioner guidance.}
Always run the tournament.
The winner is also a \emph{diagnostic}: LightGBM winning indicates a small
or identifier-heavy dataset; DeepMatcher winning indicates soft similarity
matters; GAT winning indicates a large, densely co-referent dataset (rare
at scale).
None of these conditions can be reliably predicted from schema inspection
alone---only the data reveals which signal type dominates.

\section{Lesson 2: Precision and Recall Break at Different Stages}
\label{sec:lesson2}

\noindent\textbf{Claim.} Precision and recall fail at structurally different
points in the pipeline, and the common instinct---tune the matcher
threshold---cannot fix either.

\subsection{Root causes}

\paragraph{Recall is lost before the matcher runs.}
A pair never retrieved is lost forever---no threshold adjustment recovers it.
Two retrieval failures dominate.
\emph{(i) Embedding retrievers miss surface variants.} Embedding similarity
collapses ``J.~Smith'' and ``John Smith,'' but OCR artifacts
(``Heuslein''/``Hauslein'') or heavy abbreviations push genuine matches
apart; HNSW's $M$ parameter leaves coverage gaps that compound at scale.
\emph{(ii) Embedding retrievers operate in a single modality.}
Exact categorical identifiers and structured codes produce no useful
gradient in the embedding space; two records sharing an identical
identifier but with variant text are never nominated.

\paragraph{Precision fails because sparse records look like everything.}
A record with only one populated field has nothing to disagree with; it
scores high against every other record sharing that field.
A sparse record becomes a \emph{bridge}: it matches above threshold against
two unrelated dense clusters, and transitive closure chains them into one.
This is not a matcher bug; it is the geometry of the problem.

\subsection{Fix: separate tools per stage}

\paragraph{For recall: diversify the retriever.}
We compose two structurally different retrieval strategies and union their
outputs.
\emph{Strategy 1---HNSW ensemble for embedding-space coverage}: an ensemble
of HNSW indices with diverse $(M, \mathit{ef\_search})$ configurations,
\begin{equation}
\mathcal{C}_{\text{ens}} = \bigcup_{i=1}^{N}\, \text{KNN}_k(\mathbf{E};\, M_i, \mathit{ef}_i).
\label{eq:ensemble}
\end{equation}
On MusicBrainz, a single $M{=}16$ index misses 67 true pairs (97.7\% recall);
the ensemble recovers 11 of them ($+0.4$\,pp).
\emph{Strategy 2---Identifier-based blocking for non-semantic matches}: a
lightweight exact-match inverted index over identifier fields,
$\mathcal{C}_{\text{final}} = \mathcal{C}_{\text{HNSW}} \cup \mathcal{C}_{\text{ID}}$.
On NCV, identifier blocking recovers 7 true-positive pairs the HNSW ensemble
missed entirely ($+0.3$\,pp); on DBLP-Scholar (no identifier fields), it
contributes nothing---each strategy activates only where needed
(Table~\ref{tab:hnsw_ablation}).

\begin{table}[t]
\centering
\small
\caption{Blocker recall (\%) at $k{=}20$.
$^\dagger$No identifier fields; +ID contributes 0 additional pairs.}
\label{tab:hnsw_ablation}
\setlength{\tabcolsep}{4pt}
\begin{tabular}{@{}lccc@{}}
\toprule
\textbf{Dataset} & \textbf{Single} & \textbf{Ensemble} & \textbf{+ID Blk.} \\
\midrule
DBLP-Scholar & 100.0 & 100.0 & 100.0$^\dagger$ \\
MB 200K      &  97.7 &  98.1 &  98.1$^\dagger$ \\
NCV          &  97.7 &  97.8 &  98.1 \\
\bottomrule
\end{tabular}
\end{table}

\paragraph{For precision: hard rules on top of soft classifiers.}
A learned matcher is a function of its training distribution; production
data drifts.
A model that achieved 99\% pairwise precision on validation can degrade
when field-population rates shift---and in ER the cost is not a noisy
prediction but a \emph{permanently merged cluster} that downstream
consumers inherit.
Customers also treat certain rules as non-negotiable (``different phone
number means different restaurant''), and no amount of retraining can
guarantee a soft classifier will never violate them.
We layer three deterministic guardrails on top of the matcher.

\emph{(1) Sparsity-aware thresholds.}
Training data is typically balanced by entity size, but production has a
long tail of sparse records with one or two populated fields.
A global threshold over-accepts these pairs.
We bin candidate pairs by the number of shared populated fields and learn
a separate threshold per bin, with monotonicity enforced
($\theta_b \geq \theta_{b+1}$): sparser pairs require higher confidence.
This improves purity by $+8.4$\,pp on MusicBrainz and $+1.5$\,pp on NCV;
no-op on Restaurants (full fields).

\emph{(2) Per-field hard vetoes.}
For identifier fields, a hard rule zeroes the match probability when both
records have the field populated but similarity falls below a
field-specific floor:
\begin{equation}
\hat{y} = 0 \;\;\text{if}\;\; \exists f \in \mathcal{V}\!:\; \text{both\_have}_f \wedge \text{sim}_f < \tau_f.
\label{eq:hard_veto}
\end{equation}
This improves purity by $+3.6$\,pp on MusicBrainz and $+32.8$\,pp on
Restaurants---datasets where identifier conflicts between genuinely
different entities are common.
Fields designated \texttt{no\_override} encode unconditional business rules
that no learned model can bypass.

\emph{(3) Evidence and fast-path gates.}
Two additional cheap gates compose with the sparsity threshold and veto:
an \emph{evidence} gate rejects pairs that share too few populated fields
for any matcher to be reliable, and a \emph{fast-path} gate short-circuits
pairs with very high confidence ($\geq 99.5\%$) that no hard rule
contradicts.

\begin{table}[t]
\centering
\small
\caption{Precision ablation: cluster purity (\%) as mechanisms are added
incrementally.
Baseline=tournament-winning matcher with global threshold.
Bold=best per dataset.}
\label{tab:precision_ablation}
\begin{tabular}{@{}lccc@{}}
\toprule
\textbf{Dataset} & \textbf{Baseline} & \textbf{+Sparsity} & \textbf{+Vetoes} \\
\midrule
Restaurants & 51.6 & 53.1 & \textbf{84.4} \\
Cora        & 95.9 & 98.0 & \textbf{98.1} \\
Geo Settl.  & 98.3 & \textbf{98.7} & 98.5 \\
DBLP-Sch.   & 95.4 & \textbf{99.8} & 99.5 \\
MB 200K     & 77.7 & 86.1 & \textbf{89.7} \\
NCV         & 97.4 & \textbf{98.9} & 98.6 \\
\bottomrule
\end{tabular}
\end{table}

\paragraph{Practitioner guidance.}
Diagnose before tuning.
If your largest clusters contain records from different entities: precision
problem---add field vetoes, tighten sparsity thresholds.
If singleton clusters should have been merged: recall problem---increase
HNSW $M$, add ensemble indices, check blocking coverage.
Fix retrieval gaps at the retrieval layer; lowering the matcher threshold
cannot recover pairs the retriever never nominated.

\section{Lesson 3: One False Positive Can Collapse Your Clusters}
\label{sec:lesson3}

\noindent\textbf{Claim.} Connected components (CC) clustering---the
standard post-matching step in ER~\cite{christophides2020end,
papadakis2021almost}---works well when matchers are well-calibrated but
fails when they are not.
Center-based clustering avoids error propagation but under-merges.
Active cross-cluster verification recovers recall without cascading false
merges.

\paragraph{Transitivity is an assumption, not a guarantee.}
ML matchers are not inherently
transitive~\cite{baas2021transitivity,transclean2025}: a matcher may
declare $\langle r_i, r_j\rangle$ and $\langle r_j, r_k\rangle$ as matches
while $\langle r_i, r_k\rangle$ is a non-match---a logically inconsistent
triple that CC resolves by merging all three.
When the matcher's false-positive rate is non-trivial, a single borderline
edge propagates through Union-Find and chains unrelated clusters into
mega-clusters.
Figure~\ref{fig:megacluster} shows the cascade on three ``Sakura Sushi''
records: the matcher never directly scores $r_1\!\leftrightarrow\!r_3$, yet
CC merges Portland and Seattle into one cluster.

\begin{figure}[t]
\centering
\small
\renewcommand{\arraystretch}{1.2}
\setlength{\tabcolsep}{3pt}
\begin{tabular}{@{}lllllc@{}}
\toprule
\textbf{ID} & \textbf{Name} & \textbf{Phone} & \textbf{Addr} & \textbf{City} & \textbf{True} \\
\midrule
\rowcolor{blue!15}
R1 & Sakura Sushi & 503-0147 & 42 Oak St & Portland & A \\
\rowcolor{yellow!40}
R2 & Sakura Sushi & ---      & ---       & ---      & A \\
\rowcolor{red!12}
R3 & Sakura Sushi & 206-9283 & 8 Elm Ave & Seattle  & B \\
\bottomrule
\end{tabular}

\vspace{4pt}

\scriptsize
\begin{tabular}{@{}l c c@{}}
\toprule
\textbf{Pair} & \textbf{Score} & \textbf{CC outcome} \\
\midrule
R1~$\leftrightarrow$~R2 & 0.91 & \textcolor{green!50!black}{Merged (correct)} \\
R2~$\leftrightarrow$~R3 & 0.88 & \textcolor{green!50!black}{Merged (correct)} \\
R1~$\leftrightarrow$~R3 & ---  & \textcolor{red!70}{\textbf{Never scored \(\Rightarrow\) FP}} \\
\bottomrule
\end{tabular}
\caption{Transitive closure failure on a 3-record subset of
Figure~\ref{fig:bridge}. The sparse bridge R2 lets CC chain the Portland
and Seattle entities without ever directly comparing R1 to R3. Verified
merge (\S\ref{sec:verified_merge}) forces the missing comparison and
blocks the merge.}
\label{fig:megacluster}
\end{figure}

\paragraph{Verified merge.}
\label{sec:verified_merge}
We replace blind transitive closure with a two-stage procedure.
\emph{Stage 1 (Center assignment):} each record joins the cluster of its
single highest-scoring neighbor above threshold---no edges propagate.
\emph{Stage 2 (Verified merge):} for each Stage-1 cluster pair connected
by at least one above-threshold edge in the original candidate set,
(a)~select up to $k{=}3$ representatives per cluster closest to the
centroid; (b)~score all cross-cluster representative pairs through the
matcher \emph{with hard vetoes enabled}---generating direct pairwise
evidence the blocking stage may never have produced; (c)~if \emph{any}
cross-cluster pair triggers a veto or scores below threshold, block the
merge.
A single piece of negative evidence is sufficient---this asymmetry prevents
error propagation.
For connected components of 3+ Stage-1 clusters, we verify all
$\binom{n}{2}$ cluster pairs independently to prevent transitivity from
re-entering through the merge pass itself.

\begin{table*}[t]
\centering
\small
\caption{Clustering ablation. Pair-F1, Adjusted Rand Index (ARI), and
pairwise precision per dataset.
\emph{Baseline}=center-based clustering~\cite{center_clustering}.
\emph{+Transitivity}=connected components~\cite{christophides2020end}.
\emph{+Verification}=verified merge. Bold=best per dataset.}
\label{tab:clustering}
\setlength{\tabcolsep}{5pt}
\begin{tabular}{@{}l ccc ccc ccc@{}}
\toprule
& \multicolumn{3}{c}{\textbf{Baseline}} & \multicolumn{3}{c}{\textbf{+Transitivity}} & \multicolumn{3}{c}{\textbf{+Verification}} \\
\cmidrule(lr){2-4} \cmidrule(lr){5-7} \cmidrule(lr){8-10}
\textbf{Dataset} & F1 & ARI & Prec. & F1 & ARI & Prec. & F1 & ARI & Prec. \\
\midrule
Restaurants & \textbf{1.000} & \textbf{.900} & 1.000 & \textbf{1.000} & \textbf{.900} & 1.000 & \textbf{1.000} & \textbf{.900} & 1.000 \\
Cora        & .321 & .286 & .986 & \textbf{.885} & \textbf{.853} & .914 & .876 & .779 & .994 \\
Geo Settl.  & .889 & .804 & .994 & .771 & .831 & .720 & \textbf{.972} & \textbf{.956} & .989 \\
DBLP        & .424 & .426 & .411 & .371 & .464 & .312 & \textbf{.477} & \textbf{.475} & .444 \\
MB 200K     & \textbf{.540} & \textbf{.577} & .509 & .000 & .000 & .000 & .277 & .356 & .229 \\
NCV         & \textbf{.667} & \textbf{.668} & .500 & .002 & .002 & .001 & \textbf{.667} & .647 & .500 \\
\bottomrule
\end{tabular}
\end{table*}

\paragraph{Reading the ablation.}
Table~\ref{tab:clustering} shows the three regimes.
On \emph{Restaurants} (clean, small) all three algorithms reach
Pair-F1\,$=\,1.0$: the matcher is so well-calibrated that transitivity adds
no false links.
On \emph{Cora} (99.7\% validation precision) transitivity provides the
largest gain---0.321~$\to$~0.885 F1---by recovering multi-hop links the
baseline fragments; verification is comparable (0.876) with higher
precision.
On \emph{Geo Settlements} verification dominates: F1
0.889~$\to$~\textbf{0.972}, while raw transitivity \emph{hurts}
(0.889~$\to$~0.771) as geographically similar but distinct settlements get
chained.
On \emph{MB 200K} and \emph{NCV} transitivity catastrophically collapses
F1 (0.540~$\to$~0.000 and 0.667~$\to$~0.002 respectively), as common field
values chain unrelated records into mega-clusters.
Verification preserves the baseline on NCV (0.667) and reduces damage on
MB 200K (0.277): when the underlying matcher's precision is too low
(0.509), even verification gates cannot save it.

\paragraph{The lesson.}
Transitivity is not a free lunch.
It helps when the matcher is well-calibrated (Cora, Restaurants) and
catastrophically hurts when the false-positive rate is high (MB 200K,
NCV).
The key predictor is \emph{baseline precision}: above 0.9 transitivity is
safe; below 0.5 it creates mega-clusters.
Verified merge provides a safety net across all regimes and produces the
best Pair-F1 on 4/6 datasets (strictly best on Geo Settlements and DBLP,
tied for best on Restaurants and NCV).
Its cost is modest: $O(k^2)$ additional matcher calls per candidate
cluster pair, with $k{=}3$ by default.

\section{Related Work}
\label{sec:related}

\paragraph{Classical and PLM-based ER.}
The field traces from Fellegi-Sunter~\cite{fellegi1969theory} through
Magellan~\cite{konda2016magellan} to PLM-based systems.
Ditto~\cite{li2020deep} achieved 29\% $F_1$ improvement via BERT
fine-tuning; Paganelli \etal~\cite{paganelli2024analyzing} analyzed how BERT
representations serve entity matching, and
ZeroER~\cite{wu2020zeroer} extends the paradigm to the unsupervised
setting. Thirumuruganathan \etal~\cite{thirumuruganathan2025heterogeneity} document a 40\%
$F_1$ drop from data heterogeneity---a direct motivator for our L1.

\paragraph{Graph-based ER.}
HierGAT~\cite{yao2022entity} and GraphER~\cite{hu2025grapher} encode
relational structure with attention and differential dependencies
respectively; Saeedi \etal~\cite{saeedi2025graphcr_new} use graph metrics to drive
cluster repair with active LLM feedback. These methods operate inside the
matcher and so do not address the cross-stage cascades exposed by L2 and
L3.

\paragraph{LLM-based ER and distillation.}
Peeters and Bizer~\cite{peeters2025entity} report GPT-4 outperforms transferred PLMs by
40--68\% zero-shot; Fan \etal~\cite{fan2024cost} and Wang \etal~\cite{wang2025match}
investigate cost-effective and selection-based ER paradigms.
Wadhwa \etal~\cite{wadhwa2024learning} and Steiner \etal~\cite{steiner2024fine} distill LLM reasoning
into smaller open-weight matchers, similar in spirit to our teacher /
student split. Our work differs in conditioning the teacher on an
inspectable SOP and in adding hard-rule safeguards (\S\ref{sec:lesson2})
and verified merge (\S\ref{sec:lesson3}) that the teacher itself does not
perform.

\paragraph{SOP-driven agents.}
Agent-S~\cite{li2025agents} automates SOP execution;
SOP-Bench~\cite{yin2025sopbench} shows even GPT-4o achieves 30--70\% on
complex SOPs. We are, to our knowledge, the first to apply SOP-driven
automation to entity resolution.

\section{Limitations}
\label{sec:limitations}

Our study has a few limitations that bound the scope of its conclusions.

\textbf{Public datasets by design.}
We deliberately restrict all experiments to six public benchmarks so that
every result is fully reproducible and no proprietary or customer
information is disclosed. Accordingly, we report standard accuracy metrics
(Pair-F1, purity, and Adjusted Rand Index) rather than production outcomes
such as business impact or robustness to live distribution drift.

\textbf{Data-scarce regime.}
We cap training and validation at 10K records each to reflect self-serve
deployments where labeling requires scarce domain expertise. Some findings
are specific to this budget---for example, ``GAT never wins the
tournament'' should be read as ``under a 10K-label budget,'' since a
graph-based matcher could plausibly overtake the other families given
abundant labels.

\textbf{Limited roster and single configuration.}
The tournament covers three canonical families (DeepMatcher, LightGBM,
GAT) and excludes LLM-based or cross-encoder matchers at inference. We use
a single teacher LLM, base encoder (all-MiniLM-L6-v2), and seed (42), so we
report point estimates rather than variance or significance tests.

\textbf{Verified merge has a precision floor.}
Verified merge is a safety net, not a cure: when baseline pairwise
precision is very low (e.g.\ MusicBrainz 200K at $\approx$0.51), its gates
reduce but do not prevent cluster collapse.

\textbf{Human effort is not quantified.}
SOP construction relies on a domain expert refining an LLM-drafted
specification over a few iterations; we characterize this effort
qualitatively rather than in person-hours.

\section{Conclusion}
\label{sec:conclusion}

We deployed entity resolution at scale on six benchmarks and walked away
with three findings that we wish someone had handed us at the start.

First, no single matching algorithm wins across datasets
(\S\ref{sec:lesson1}). In a small per-dataset bake-off across three
canonical families, DeepMatcher and LightGBM each took the top spot on
three benchmarks and GAT never won. Any team that commits to one
architecture therefore loses on at least a third of the datasets it has
yet to see, and the bake-off pays for itself many times over.

Second, precision and recall fail at different stages and need different
mechanisms (\S\ref{sec:lesson2}). Moving the matcher score threshold
trades one off against the other and solves neither. We improve
precision only by adding hard rule-based vetoes that the matcher cannot
learn from data, and we improve recall only by running several diverse
blocking strategies in parallel.

Third, one false-positive link can silently merge unrelated entity groups
because transitive closure compounds matcher errors
(\S\ref{sec:lesson3}). We repeatedly saw single low-evidence links chain
hundreds of records that shared nothing in common into one giant
cluster. Our verified-merge step---which re-runs the matcher on
representative pairs across every candidate merge before committing---
produces the best Pair-F1 on four of the six datasets and avoids the
catastrophic precision collapse that transitive closure inflicts on the
two largest.

None of these three mechanisms is individually novel, but adopting all
three as defaults rather than as escalation paths is what changed our
production error profile. We hope the recipes here give other teams a
shorter route to the same outcome.

\bibliographystyle{IEEEtran}
\bibliography{references}

\appendix

\section{Framework Details}
\label{app:framework}

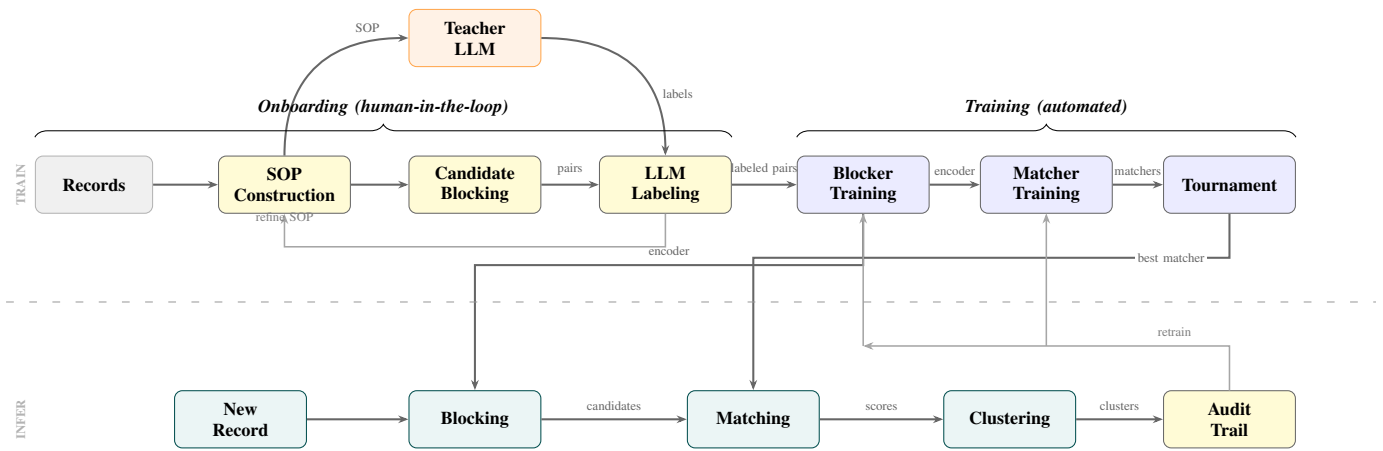
\begin{figure*}[t]
\centering
\resizebox{\textwidth}{!}{%
\begin{tikzpicture}[
    box/.style={rectangle, draw=black!60, fill=blue!8, minimum width=1.8cm,
        minimum height=0.78cm, align=center, rounded corners=3pt, font=\scriptsize\bfseries},
    hbox/.style={rectangle, draw=black!60, fill=yellow!20, minimum width=1.8cm,
        minimum height=0.78cm, align=center, rounded corners=3pt, font=\scriptsize\bfseries},
    tbox/.style={rectangle, draw=orange!70, fill=orange!10, minimum width=1.8cm,
        minimum height=0.78cm, align=center, rounded corners=3pt, font=\scriptsize\bfseries},
    ibox/.style={rectangle, draw=teal!60!black, fill=teal!8, minimum width=1.8cm,
        minimum height=0.78cm, align=center, rounded corners=3pt, font=\scriptsize\bfseries},
    gbox/.style={rectangle, draw=gray!60, fill=gray!12, minimum width=1.6cm,
        minimum height=0.78cm, align=center, rounded corners=3pt, font=\scriptsize\bfseries},
    arr/.style={-{Stealth[length=4pt]}, thick, black!60},
    feedback/.style={-{Stealth[length=4pt]}, black!35, semithick},
]
\node[gbox] (recs)    at (-2.0,  0) {Records};
\node[hbox] (sop)     at (0.6,   0) {SOP\\Construction};
\node[hbox] (knnb)    at (3.2,   0) {Candidate\\Blocking};
\node[hbox] (label)   at (5.8,   0) {LLM\\Labeling};
\node[box]  (siam)    at (8.5,   0) {Blocker\\Training};
\node[box]  (train)   at (11.0,  0) {Matcher\\Training};
\node[box]  (tourn)   at (13.5,  0) {Tournament};
\node[tbox] (teacher) at (3.2, 2.0) {Teacher\\LLM};
\node[ibox] (qrec)    at (0.0,  -3.2) {New\\Record};
\node[ibox] (knn)     at (3.2,  -3.2) {Blocking};
\node[ibox] (score)   at (7.0,  -3.2) {Matching};
\node[ibox] (cluster) at (10.5, -3.2) {Clustering};
\node[hbox] (audit)   at (13.5, -3.2) {Audit\\Trail};
\draw[arr] (recs.east)  -- (sop.west);
\draw[arr] (sop.east)   -- (knnb.west);
\draw[arr] (knnb.east)  -- (label.west)  node[above, midway, font=\tiny] {pairs};
\draw[arr] (label.east) -- (siam.west)   node[above, midway, font=\tiny] {labeled pairs};
\draw[arr] (siam.east)  -- (train.west)  node[above, midway, font=\tiny] {encoder};
\draw[arr] (train.east) -- (tourn.west)  node[above, midway, font=\tiny] {matchers};
\draw[arr] (sop.north)    .. controls +(0,1.4) and +(-0.7,0) .. (teacher.west)
    node[above, near end, font=\tiny] {SOP};
\draw[arr] (teacher.east) .. controls +(0.7,0) and +(0,1.4)  .. (label.north)
    node[right, near end, font=\tiny] {labels};
\draw[arr] (siam.south) -- ++(0,-0.7) -| (knn.north)
    node[pos=0.25, above, font=\tiny] {encoder};
\draw[arr] (qrec.east)    -- (knn.west);
\draw[arr] (knn.east)     -- (score.west)   node[above, midway, font=\tiny] {candidates};
\draw[arr] (score.east)   -- (cluster.west) node[above, midway, font=\tiny] {scores};
\draw[arr] (cluster.east) -- (audit.west)   node[above, midway, font=\tiny] {clusters};
\draw[arr] (tourn.south) -- ++(0,-0.6) -| (score.north)
    node[right, font=\tiny, fill=white, inner sep=1pt, pos=0.10] {best matcher};
\draw[feedback] (label.south) -- ++(0,-0.45) -| (sop.south)
    node[pos=0.75, above, font=\tiny, text=black!50] {refine SOP};
\draw[feedback]
    (13.5, -2.81) -- (13.5, -2.2)
    -- (8.5, -2.2)
    node[pos=0.15, above, font=\tiny, text=black!50] {retrain};
\draw[feedback] (11.0, -2.2) -- (11.0, -0.39);
\draw[feedback] (8.5, -2.2) -- (8.5, -0.39);
\draw[loosely dashed, black!40, thin] (-3.2,-1.6) -- (15.5,-1.6);
\node[rotate=90, font=\tiny\bfseries, text=black!30] at (-3.0,  0.0) {TRAIN};
\node[rotate=90, font=\tiny\bfseries, text=black!30] at (-3.0, -3.2) {INFER};
\draw[decoration={brace, amplitude=5pt}, decorate]
    ($(recs.north west)+(0,0.25)$) -- ($(label.north east)+(0,0.25)$)
    node[midway, above=5pt, font=\scriptsize\bfseries\itshape]
        {Onboarding (human-in-the-loop)};
\draw[decoration={brace, amplitude=5pt}, decorate]
    ($(siam.north west)+(0,0.25)$) -- ($(tourn.north east)+(0,0.25)$)
    node[midway, above=5pt, font=\scriptsize\bfseries\itshape]
        {Training (automated)};
\end{tikzpicture}%
}
\caption{Full ER pipeline. LLM-driven stages (yellow, orange) involve
humans; cost-sensitive matching/clustering (teal) run on lightweight
distilled models. Solid arrows: data flow. Faded arrows: feedback loops
(SOP refinement and matcher retraining).}
\label{fig:arch}
\end{figure*}

Figure~\ref{fig:arch} shows the full pipeline: \emph{onboarding}
(SOP construction, blocking, LLM labeling---all human-in-the-loop, top
row), \emph{training} (blocker and matcher distillation plus tournament,
top row right), and \emph{inference} (blocking, matching, clustering,
audit---bottom row).
LLMs touch only the human-facing stages; matching and clustering run
entirely on lightweight distilled models.

\paragraph{Why SOPs (extended).}
There is no universal definition of an ``entity.''
Consider two restaurant listings sharing phone, address, and city but
with different names (e.g.\ \emph{Sakura Sushi} and \emph{Thai Orchid}
at the same food-court address): a generic ER system says match (three
strong fields agree); a domain expert says no---multiple tenants share
one phone line.
The rule ``same name + same phone = match; phone alone is insufficient''
cannot be inferred from data without examples no organization possesses.
An LLM drafts an initial SOP from the schema; a domain expert refines
it in 3--5 iterations.
SOPs vary substantially across benchmarks---even two bibliographic
datasets (Cora, DBLP-Scholar) require structurally different SOPs.

\paragraph{Pipeline phases (extended).}
\textit{Onboarding (human-in-the-loop).} The SOP-conditioned LLM
teacher labels candidate pairs surfaced by approximate $k$-NN blocking,
producing per-field similarity assessments, a confidence score, and
natural-language evidence; a domain expert reviews low-confidence labels
and refines the SOP (typically 2--3 rounds).
\textit{Training (automated).} Labeled pairs train the blocker encoder
and matcher independently; the tournament evaluates all matchers on
held-out validation pairs.
\textit{Inference.} An ensemble of blockers (multiple HNSW indices plus
identifier-based blocking) generates candidates that the
tournament-winning matcher scores with safeguard layers
(\S\ref{sec:lesson2}); verified merge (\S\ref{sec:lesson3}) then clusters
records using conservative direct assignment followed by verified
cross-cluster merging.

\section{SOP Excerpt}
\label{app:sop_yaml}

Figure~\ref{fig:sop_yaml} shows a SOP excerpt for restaurant matching,
illustrating how field importance, acceptable variations, and decision
boundaries are encoded as inspectable YAML.

\begin{figure}[h]
\centering
\small
\begin{verbatim}
sop:
  version: "1.2"
  domain: restaurant_matching
  field_hierarchy:
    critical: [name, phone]
    high:     [address]
    medium:   [city, cuisine]
    low:      [zipcode]
  tolerances:
    name:
      - type: abbreviation
      - type: typo  # max_edit_dist: 2
    phone:
      - type: formatting
  decision_rules:
    match:     ">=2 critical agree, 0 conflict"
    review:    "1 conflict + >=2 high agree"
    non_match: ">=2 critical conflict"
\end{verbatim}
\caption{SOP excerpt for restaurant matching.}
\label{fig:sop_yaml}
\end{figure}

\end{document}